\documentclass[final]{cvpr}

\usepackage{times}
\usepackage{epsfig}
\usepackage{graphicx}
\usepackage{amsmath}
\usepackage{amssymb}

\newcommand\blfootnote[1]{%
  \begingroup
  \renewcommand\thefootnote{}\footnote{#1}%
  \addtocounter{footnote}{-1}%
  \endgroup
}

\usepackage{dsfont}
\newcommand{\concat}{%
  \mathbin{{+}\mspace{-8mu}{+}}%
}
\usepackage[bottom]{footmisc}
\usepackage{textcomp}
\usepackage[dvipsnames]{xcolor}
\definecolor{mygray}{gray}{0.5}
\definecolor{mygreen}{rgb}{0.0, 0.42, 0.24}
\definecolor{myred}{rgb}{1.0, 0.25, 0.25}
\definecolor{myreal}{rgb}{0.96, 0.49, 0.49}
\definecolor{myfake}{rgb}{0.56, 0.67, 0.86}
\definecolor{mylayer}{rgb}{0.4, 0.26, 0.13}

\definecolor{cora}{rgb}{1,0,0}
\definecolor{citeseer}{rgb}{0,0.5,0}
\definecolor{pubmed}{rgb}{0,0,1}
\definecolor{tox21}{rgb}{0,0.75,0.75}
\definecolor{bbbp}{rgb}{0.75,0,0.75}

\usepackage{setspace}
\usepackage{cancel}
\usepackage{multirow,multicol}
\usepackage{array}
\newcolumntype{?}{!{\vrule width 1pt}}

\usepackage[pagebackref=true,breaklinks=true,colorlinks,bookmarks=false]{hyperref}

\def\cvprPaperID{1368} 
\def\confYear{CVPR 2021}

\begin{document}

\title{Exploiting Edge-Oriented Reasoning for 3D Point-based Scene Graph Analysis}

\author{Chaoyi Zhang\\
University of Sydney\\
{\tt\small chaoyi.zhang@sydney.edu.au}
\and
Jianhui Yu\\
University of Sydney\\
{\tt\small jianhui.yu@sydney.edu.au}
\and
Yang Song\\
University of New South Wales\\
{\tt\small yang.song1@unsw.edu.au}
\and
Weidong Cai\\
University of Sydney\\
{\tt\small tom.cai@sydney.edu.au}
}

\maketitle

\begin{abstract}
Scene understanding is a critical problem in computer vision. 
In this paper, we propose a 3D point-based scene graph generation ($\mathbf{SGG_{point}}$) framework to effectively bridge perception and reasoning to achieve scene understanding via three sequential stages, namely scene graph construction, reasoning, and inference.
Within the reasoning stage, an EDGE-oriented Graph Convolutional Network (\texttt{EdgeGCN}) is created to exploit multi-dimensional edge features for explicit relationship modeling, together with the exploration of two associated twinning interaction mechanisms between nodes and edges for the independent evolution of scene graph representations.
Overall, our integrated $\mathbf{SGG_{point}}$ framework is established to seek and infer scene structures of interest from both real-world and synthetic 3D point-based scenes. 
Our experimental results show promising edge-oriented reasoning effects on scene graph generation studies. We also demonstrate our method advantage on several traditional graph representation learning benchmark datasets, including the node-wise classification on citation networks and whole-graph recognition problems for molecular analysis.\blfootnote{Our project page: \href{https://sggpoint.github.io/}{$\mathbf{SGG_{point}}$.github.io}}
\end{abstract}

\section{Introduction}
Scene understanding is intrinsically close to the essence of computer vision. 
It simulates human visual system in recognizing the miscellaneous clues concealed in the complex visual world, succeeded by understanding what we perceive in the visual scenes surrounding us~\cite{fei2004we}. 
This process could be integrated and assisted with an efficient use of semantic scene graph ($\mathbf{SG}$), which has its popularity well-demonstrated within the computer graphics community, via depicting the objects and their inner structural relationships (scene layouts) as its nodes and edges, respectively.

Unlike most of the successful works proposed for 2D $\mathbf{SG}$ studies~\cite{GraphRCNN,IterMessPassing,visualpriors,ImgGeneration_CVPR18}, this paper focuses on 3D point-based semantic $\mathbf{SG}$ analysis -- an emerging 3D visual recognition task that has not been well-explored yet. 
Such methods could provide great aid for arising cross-domain vision tasks including 2D-3D scene retrieval~\cite{3DSSG2020}, 3D visual grounding~\cite{chen2020scanrefer, ReferIt3D}, and scene captioning~\cite{chen2020scan2cap}, which would subsequently benefit real-life applications such as creative interior decoration designs, self-driving autonomous vehicles, or other AI-enriched indoor/outdoor industries.

\begin{figure}[t]
    \label{image:first}
    \centering
    \includegraphics[width=0.5\textwidth]{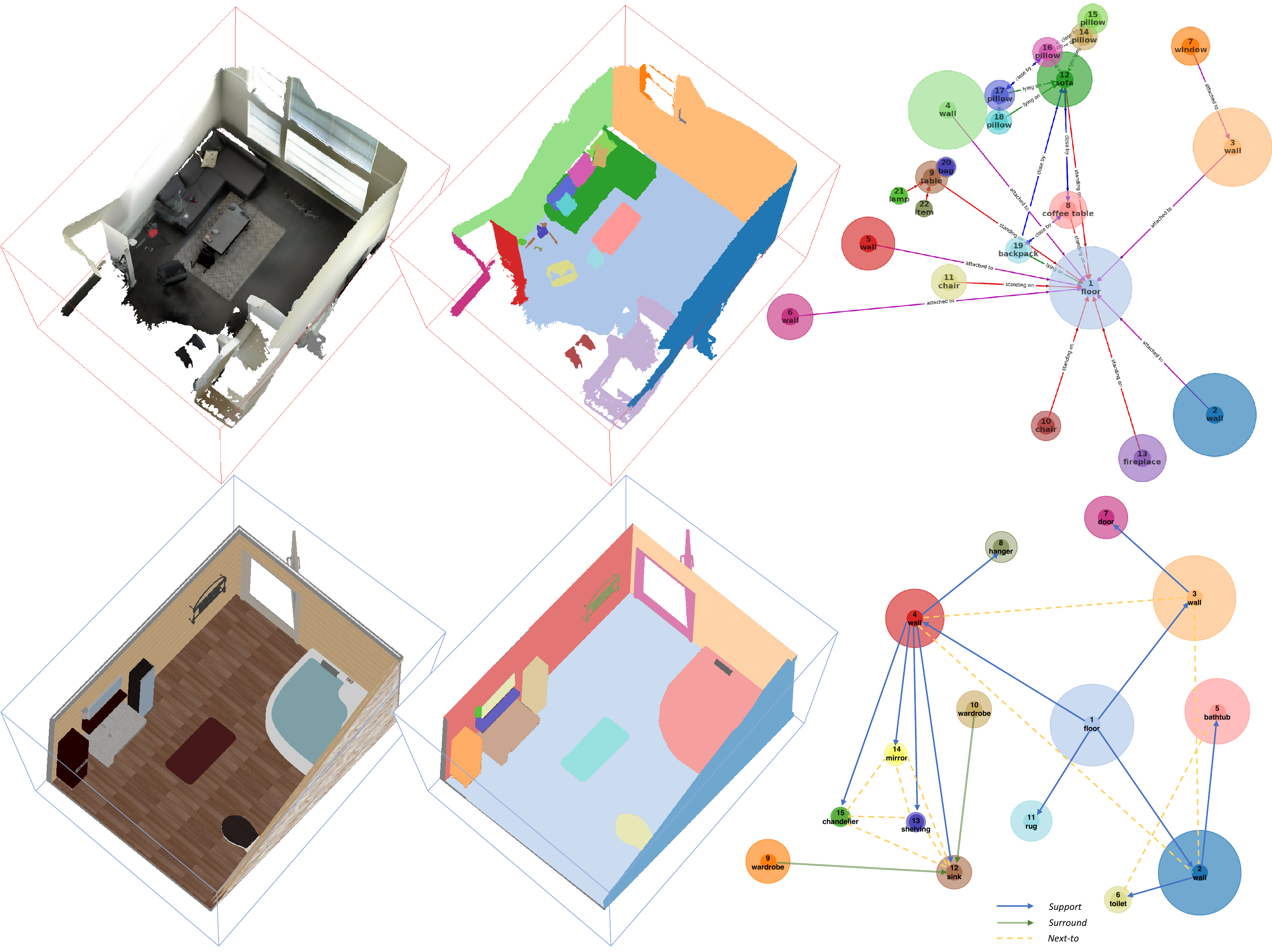}
    \caption{3D Point-based Scene Graph Generation ($\mathbf{SGG_{point}}$) takes as inputs \textcolor{myreal}{\textbf{real-world}} or \textcolor{myfake}{\textbf{synthetic}} 3D scenes $\mathcal{S}$ (left) and class-agnostic instance mask $\mathcal{M}$ (middle) to inference a scene graph $\mathcal{G}$ (right). $\mathcal{M}$ and $\mathcal{G}$ above are aligned to the same spatial layout, sharing a unified instance color encoding.}

\end{figure}

Within the rising progression of 3D point-based semantic $\mathbf{SG}$ analysis, the research interests have gradually shifted from object-centric point cloud learning tasks, such as 3D object detection~\cite{PointRCNN, pixor,pointgnn}, instance segmentation~\cite{pointgroup,JSNET,asis}, and semantic scene segmentation~\cite{randla, Xu_2020_CVPR}, to the joint recognition of both objects and inter-object structural relationships, which could be further regressed to generate $\mathbf{SG}$s describing some desired scene layouts (Fig.~\ref{image:first}) for given point-based 3D scenes. 
Moreover, existing works~\cite{GraphRCNN,3DSSG2020} have mostly treated the inter-object structural relationships as by-products derived from graph node recognition, losing sight of the visual cues lurking inside each $\mathbf{SG}$ representation and thus degrading their joint recognition performance.

In this paper, we propose a 3D $\mathbf{SGG_{point}}$ framework capable of effectively bridging perception and reasoning to achieve 3D scene understanding through three sequential stages, namely scene graph construction, reasoning, and inference. The contributions of this paper are summarized as follows:
1) To endow the graph convolution networks (GCNs) with edge-assisted reasoning capability, an edge-oriented GCN (\texttt{EdgeGCN}) is proposed to exploit multi-dimensional edge features for explicit inter-node relationship modeling. 
2) Two twinning interactions between $\mathbf{SG}$ nodes and edges are further explored to conduct comprehensive $\mathbf{SG}$ reasoning for each individual $\mathbf{SG}$ representation evolution, so that the node- and edge-oriented visual clues can be better perceived and utilized to assist the other ones' evolution via an attentional manner.
3) Our integrated $\mathbf{SGG_{point}}$ framework is demonstrated to be handy for generating 3D scene structures from either computer-aided 3D scene synthesis or real-world 3D scans, while our edge-driven interaction scheme is also proven beneficial to conventional graph representation learning tasks.

\section{Related Work}
\noindent\textbf{2D scene graph analysis.}
$\mathbf{SG}$s were firstly introduced into computer vision to capture more semantic information of objects and their inter-relationships for image retrieval~\cite{ImgRetrievalSG}. 
Thereafter, a string of image-based $\mathbf{SG}$ generation methods~\cite{IterMessPassing,motifs,li2017msdn,GraphRCNN,li2018fnet,pix2graphs} was substantially fostered by the release of the Visual Genome~\cite{visualgenome} dataset, which includes large-scale $\mathbf{SG}$ annotations on images. Xu~\etal~\cite{IterMessPassing} adopted gated recurrent units (GRUs)~\cite{GRU} to propagate messages iteratively between the primal and dual graphs formed by $\mathbf{SG}$ nodes and edges, while MotifNet~\cite{motifs} generated $\mathbf{SG}$s from global context parsed through bidirectional LSTM~\cite{LSTM}. 
Most methods~\cite{li2017msdn,li2018fnet,pix2graphs} tackled $\mathbf{SG}$ prediction problem within an object detector-cored framework for node- and edge-specific feature extraction, whereas Graph R-CNN~\cite{GraphRCNN} proposed an attentional variant of GCN~\cite{GCN} and combined it with Faster R-CNN~\cite{faster} to process contextual information between objects and relationships.
Unlike most of them that treat edge features as by-products derived from the 2D object recognition progress, we address this issue by handling nodes and edges equivalently and simultaneously as a pair of twining representations among 3D point-based scenes.

\noindent\textbf{3D point-based scene understanding.}
Differing from voxelization-based~\cite{voxnet} or view-based approaches~\cite{multiview,multiview2}, point cloud processing techniques have been advanced to support direct point-based manipulations on 3D objects or scenes~\cite{pointnet,pointnet2, dgcnn}.
They transformed 3D scene understanding into several object-centric recognition tasks including semantic scene segmentation~\cite{randla, Xu_2020_CVPR}, scene instance segmentation~\cite{pointgroup,JSNET,asis}, and scene object detection~\cite{PointRCNN, pixor,pointgnn}, which ensures the deep learning advances could be inherited from 2D vision to enhance 3D object-oriented recognition performance. 
Additionally, other concurrent 3D scene understanding works have compiled a few augmented reality focused applications such as indoor scene synthesis and augmentation~\cite{li2018grains,Kai1,kai2,zhou2019scenegraphnet}, by producing object recommendation lists for given query positions within 3D class-known scenes.
GRAINS~\cite{li2018grains} adopted recursive auto-encoders for semantic scene completion over the $\mathbf{SG}$s being organized in tree structures, while SceneGraphNet~\cite{zhou2019scenegraphnet} achieved iterative scene synthesis by passing relationship-specific messages among $\mathbf{SG}$ nodes.
Apart from these investigations in object-oriented scene recognition, only a few works have spotlighted 3D scene-oriented reasoning and understanding, by encoding the scene layouts of interest or regressing the inter-object structural relationships, due to the lack of 3D $\mathbf{SG}$ datasets.
Recently, the 3RScan~\cite{Wald2019RIO} dataset, which had been initially probed for 3D object instance re-localization task, was later upgraded as a newly established benchmark~\cite{3DSSG2020} for learning 3D semantic $\mathbf{SG}$s from point-based indoor environments.
In this work, we selected these two datasets to evaluate our approaches on 3D real-world scans.
Another synthetic dataset~\cite{song2016suncg} with scene layout annotations released in~\cite{zhou2019scenegraphnet} was also adopted for our method evaluation on 3D synthetic scenes.

\noindent\textbf{Graph-based reasoning.}
Building upon GCNs~\cite{GCN} as their core components, graph reasoning approaches conduct graph-based information propagation to achieve global relation reasoning effects among the graph nodes.
GCU~\cite{GCU} initiated a three-stage graph reasoning paradigm for 2D vision tasks with the graph projection, convolution, and re-projection operations, while GloRe~\cite{glore} and LatenGNN~\cite{LatenGNN} strengthened their global reasoning powers via flexible feature aggregations performed within their so-called interactive (or latent) space. 
Meanwhile, SGR~\cite{SGR} and GIN~\cite{GIN} inspected contextual reasoning over commonsense graph structures and utilized external knowledge to improve performance on several 2D segmentation benchmarks.
However, most existing approaches focused on relation reasoning among graph nodes, neglecting their twinning representations, i.e., graph edges.
By contrast, inspired by EGNN~\cite{EGNN}, we ameliorate GCN to make it compatible with explicit relationship modeling as desired and further exploit edge-oriented reasoning for $\mathbf{SG}$ representation learning.

\begin{figure*}[t]
    \centering
    \includegraphics[width=\textwidth]{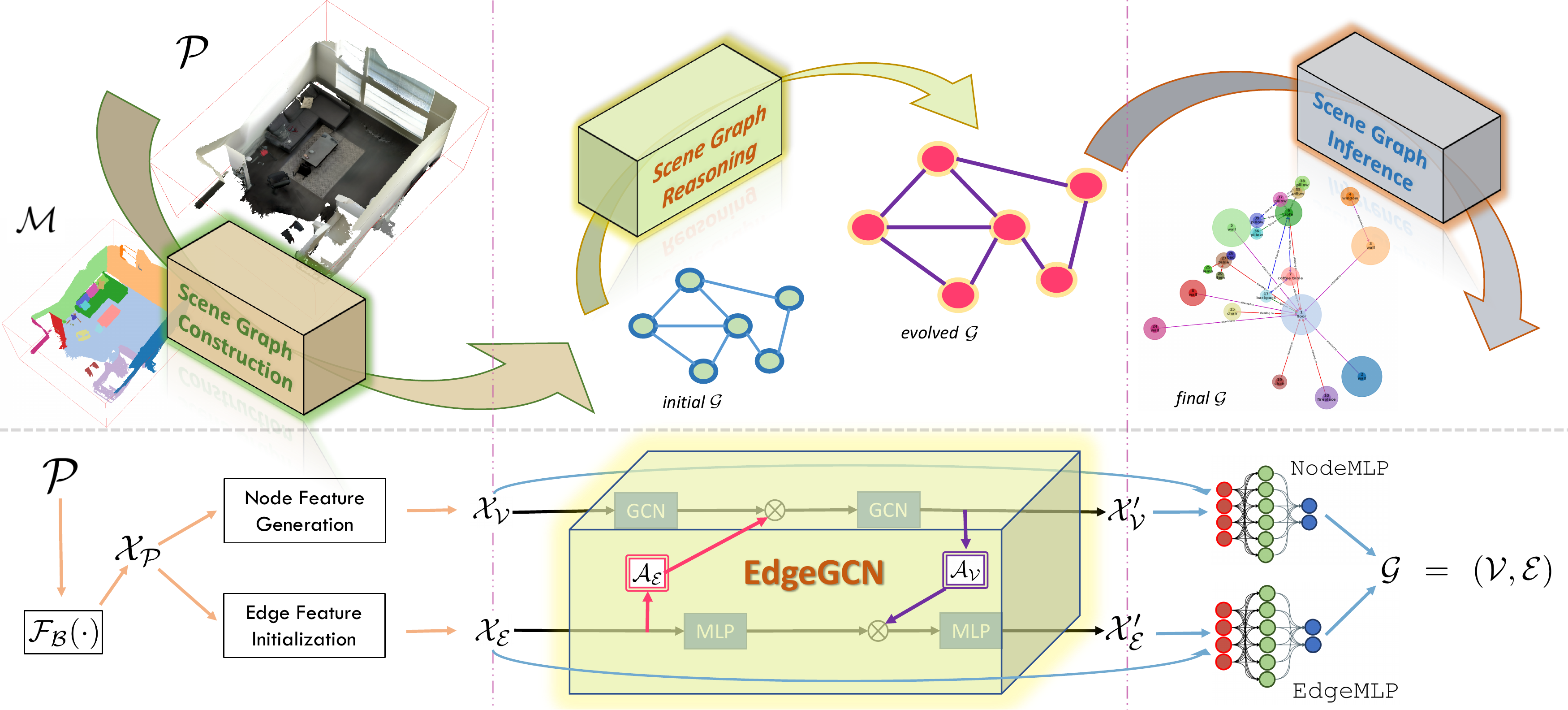}
    \caption{Our proposed 3D point-based scene graph generation ($\mathbf{SGG_{point}}$) framework consisting of three sequential stages.} 
    \label{image:pointsgg}

\end{figure*}

\section{Method}

Suppose a 3D point cloud $\mathcal{P}$ consists of $\mathcal{N}$ points $\{\mathcal{P}_k\}_{k=1,...,\mathcal{N}}$.
The ultimate goal of our point-based scene graph generation ($\mathbf{SGG_{point}}$) framework is to create a scene graph $\mathcal{G} = (\mathcal{V}, \mathcal{E})$, where nodes $\mathcal{V}$ and edges $\mathcal{E}$ depict the instance objects and their inner structural relationships, respectively. This objective can be investigated through several stages: namely scene graph construction (Construction$_{\mathbf{SG}}$ in \textit{Sec.}~\ref{sec:3.1}), reasoning (Reasoning$_{\mathbf{SG}}$ in \textit{Sec.}~\ref{sec:3.2}), and inference (Inference$_{\mathbf{SG}}$ in \textit{Sec.}~\ref{sec:3.3}).

\subsection{Scene Graph Construction}
\label{sec:3.1}
Compared to the existing work~\cite{3DSSG2020} that employed two separate backbone networks to extract independent object- and relationship-specific features, we reduce the superfluous redundancies in scene understanding via sharing one single backbone denoted as $\mathcal{F_B(\cdot)}$ to capture the point-wise features $\mathcal{X}_\mathcal{P}\in \mathcal{R}^{\mathcal{N} \times \mathcal{C}_{point}}$ from a specific $\mathcal{P} \in \mathcal{R}^{\mathcal{N} \times \mathcal{C}_{input}}$ that forms scene $S$, where $\mathcal{C}_{input}$ and $\mathcal{C}_{point}$ denote the channel numbers for inputted point clouds and their extracted point-wise features, respectively.
$\mathcal{X}_\mathcal{P}$ is further propagated to facilitate the initial modeling of the representations of ${m}$ nodes and $m^2$ one-to-one edges in $\mathcal{G}$, as $\mathcal{X}_{\mathcal{V}}\in\mathcal{R}^{m \times \mathcal{C}_{node}}$ and $\mathcal{X}_\mathcal{E}\in \mathcal{R}^{m \times m \times \mathcal{C}_{edge}}$, respectively, where $\mathcal{C}_{node}$ and $\mathcal{C}_{edge}$ indicate the channel numbers for node and edge features constructed within $\mathcal{G}$, respectively.

\noindent\textbf{Node feature generation.} 
As suggested in~\cite{3DSSG2020}, a symmetric pooling function $g(\cdot)$~\cite{pointnet} is performed on an unordered set, along the class-agnostic point-to-instance indicator $\mathcal{M}\in \{1,...,m\}^{\mathcal{N}}$ to generate the instance-wise visual signatures $\mathcal{X}_{\mathcal{V}_i}\in\mathcal{R}^{1 \times \mathcal{C}_{node}}$ for each object $i$ inside $S$, from the point-wise features $\mathcal{X}_\mathcal{P}$ obtained by $\mathcal{F}_{\mathcal{B}}(\cdot)$.
This masking operation can be formally described as:
\begin{equation}
\mathcal{X}_{\mathcal{V}_i} = g\bigg(
\Big\{
\delta(\mathcal{M}_{k}, i) \cdot \mathcal{X}_{\mathcal{P}_k}
\Big\}_{k=1,...,\mathcal{N}}
\bigg),
\end{equation}
where $\delta(\cdot,\cdot)$ denotes the Kronecker Delta. Our initial node features can now be modeled as $\mathcal{X}_{\mathcal{V}}$ by stacking together all $m$ instance-wise visual signatures across $S$.

\noindent\textbf{Edge feature initialization.} In contrast to some $\mathbf{SG}$ studies on 2D images~\cite{ImgGeneration_CVPR18} or 3D point clouds~\cite{3DSSG2020} that reformulated inter-object structural relationships as special kinds of nodes, the $\mathbf{SGG_{point}}$ framework would instead learn and encapsulate this information as multi-dimensional edge features $\mathcal{X}_\mathcal{E}$.
Each member $\mathcal{X}_{\mathcal{E}_{(i,j)}} \in\mathcal{R}^{1 \times 1 \times \mathcal{C}_{edge}}$ records the $\mathcal{C}_{edge}$-dim status of each directional connection ${\mathcal{E}_{(i,j)}}$ that points from subject $\mathcal{V}_i$ toward object $\mathcal{V}_j$, which can be initialized as $\mathcal{X}_{\mathcal{E}_{(i, j)}} = \left(\mathcal{X}_{\mathcal{V}_i} \concat
(\mathcal{X}_{\mathcal{V}_j} - \mathcal{X}_{\mathcal{V}_i})\right)$ using feature engineering and concatenation scheme introduced in~\cite{dgcnn}. 

\subsection{Scene Graph Reasoning via \textbf{\texttt{EdgeGCN}}}
\label{sec:3.2}

Previous $\mathbf{SG}$ works mostly acquired edge predictions as by-products derived from node representation learning, which might underestimate potential impacts of the visual cues lurking inside both node and edge representations toward their joint $\mathbf{SGG_{point}}$ task.
Instead, we posit that both nodes and edges are expected to be treated equally and processed simultaneously as pairs of twinning representations within a given $\mathbf{SG}$, and we thus assign each one with an exclusive learning branch and investigate graph reasoning techniques for their feature representation enhancements.

Recall the recently proposed global relation reasoning approaches~\cite{glore,SGR,GIN,GCU} that applied GCNs to perform node-wise message propagation to obtain their so-called \textit{evolved node features} through graph-based reasoning~\cite{SGR}.
We first borrow their ideas and naming rules to establish our $\mathbf{SG}$ node evolution stream, where an edge-driven interaction mechanism dubbed twinning edge attention, is proposed to enhance node-wise reasoning. Similarly, an edge evolution stream equipped with twining node attention scheme is next designed to extract node-specific cues for conducting edge-wise reasoning.

\subsubsection{Twinning Edge Attention for Node Evolution}

\begin{figure}[h]
    \centering
    \includegraphics[width=0.5\textwidth]{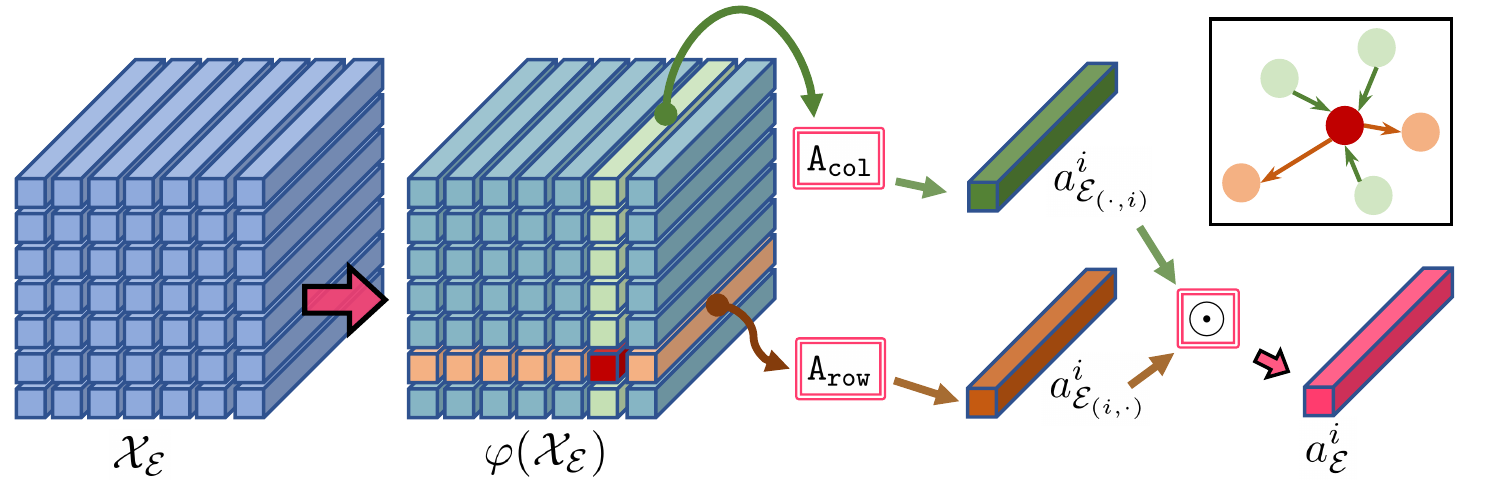}
    \caption{Twinning edge attention $\mathcal{A}_{\mathcal{E}}$ design within our \texttt{EdgeGCN} for modeling the edge-driven interactions toward node evolution.}
    \label{image:Ae}

\end{figure}

\noindent\textbf{Twinning edge attention.} Our twinning edge attention scheme is proposed to learn a multi-dimensional attention mask $\mathcal{A}_\mathcal{E} \in \mathcal{R}^{m\times\mathcal{C}^{\prime}_{node}}$ to be applied over $\mathcal{V}$ in accordance with their node-wise importance cues embedded in $\mathcal{X}_\mathcal{E}$, where $\mathcal{C}^{\prime}_{node}$ is pre-defined here to match the inner channel number within node evolution stream to be described below.
To make use of the directional status recorded in $\mathcal{X}_\mathcal{E}$, given a node $\mathcal{V}_i$, we compute its edge interaction vector $a^{i}_{\mathcal{E}} \in \mathcal{R}^{1 \times \mathcal{C}^{\prime}_{node}}$ by considering both circumstances when it plays the roles of sources or targets in various connections. 
Mathematically, the outgoing interaction signals emitted by $\mathcal{V}_i$ (as sources) and the incoming interaction signals received by $\mathcal{V}_i$ (as targets) can be captured and aggregated as $a^i_{\mathcal{E}_{(i,\cdot)}}$ and $a^i_{\mathcal{E}_{(\cdot,i)}}$, respectively, through:

\begin{equation}
a^i_{\mathcal{E}_{(i,\cdot)}} = \mathtt{A}_{\mathtt{row}} \left(
\{
W^T_{\varphi}
\mathcal{X}_{\mathcal{E}_{(i, k)}}
 \mid \forall \mathcal{V}_k
\}
\right),
\end{equation}

\begin{equation}
a^i_{\mathcal{E}_{(\cdot,i)}} = \mathtt{A}_{\mathtt{col}} \left(
\{
W^T_{\varphi}
\mathcal{X}_{\mathcal{E}_{(k, i)}}
\mid \forall \mathcal{V}_k
\}
\right),  
\end{equation}
where $W_{\varphi} \in \mathcal{R}^{\mathcal{C}_{edge}\times\mathcal{C}^{\prime}_{node}}$ is a trainable transformation matrix for converting each edge feature $\mathcal{X}_{\mathcal{E}_{(\cdot,\cdot)}} \in \mathcal{R}^{\mathcal{C}_{edge}}$ into the dimension $\mathcal{C}^{\prime}_{node}$, while $\mathtt{A_{\mathtt{row}}}(\cdot)$ and $\mathtt{A_{\mathtt{col}}(\cdot)}$ represent the channel-wise aggregation functions performed along row- and column-directions, respectively. 
Hence, as demonstrated in Fig.~\ref{image:Ae}, the overall edge-driven interaction score of node $\mathcal{V}_i$ can now be jointly learned as:
\begin{equation}
a^i_{\mathcal{E}} = \sigma (a^i_{\mathcal{E}_{(i,\cdot)}} \odot a^i_{\mathcal{E}_{(\cdot, i)}}),
\end{equation}
where $\odot$ denotes the Hadamard Product and $\sigma$ indicates the sigmoid function to emphasize the meaningful interactions and suppress the uninformative ones.

\noindent\textbf{Node evolution stream.} 
Suppose $A_{\mathcal{G}}$ as the adjacency matrix defined over $\mathcal{G}$.
Based on the definition of edge-driven twinning attention, our evolved $\mathbf{SG}$ node representation $\mathcal{X}^{\prime}_{\mathcal{V}}$ could now be learnt as:

\begin{equation} \label{eq1}
\mathcal{X}^{\prime}_{\mathcal{V}} = 
f\bigg(
\widehat{A}_{\mathcal{G}}
\Big(
\textcolor{mylayer}{
f(\widehat{A}_{\mathcal{G}}\mathcal{X}_{\mathcal{V}}W_\mathtt{G1})}
\odot \mathcal{A}_{\mathcal{E}}
\Big)
W_\mathtt{G2}
\bigg)
,
\end{equation}
where $\mathcal{A}_{\mathcal{E}}$ and $f$ denote the edge-driven interactive score and non-linear activation function, respectively, and $\widehat{A}_\mathcal{G}$ is a symmetric Laplacian matrix normalized from $A_\mathcal{G}+I$ such that all rows sum to one~\cite{GCN}, while $W_\mathtt{G1}\in\mathcal{R}^{\mathcal{C}_{node} \times\mathcal{C}^{\prime}_{node}}$ and $W_\mathtt{G2}\in\mathcal{R}^{\mathcal{C}^{\prime}_{node} \times\mathcal{C}_{node}}$ are learnable weights for two consecutive graph convolution layers to squeeze and expend the channels of node features through $\mathcal{C}_{node} \xrightarrow{} \mathcal{C}^{\prime}_{node} \xrightarrow{} \mathcal{C}_{node}$. \textbf{Note:} The inner layer outputs in Eq.~\ref{eq1} and~\ref{eq2} are indicated in \textcolor{mylayer}{\textit{dark brown}} for convenience.
Unlike~\cite{zhou2019scenegraphnet} which provides detailed relationship categorical information to guide their node evolution and employs several independent GRUs to conduct the message passing for each kind of relationship, we instead only reveal the class-agnostic relationship existences (i.e., $A_{\mathcal{G}}$) for our approach designs, leading to a flexible scalability upon the various dataset-dependent inter-object relationship annotations across datasets.

\subsubsection{Twinning Node Attention for Edge Evolution}

\begin{figure}[h]
    \centering
    \includegraphics[width=0.5\textwidth]{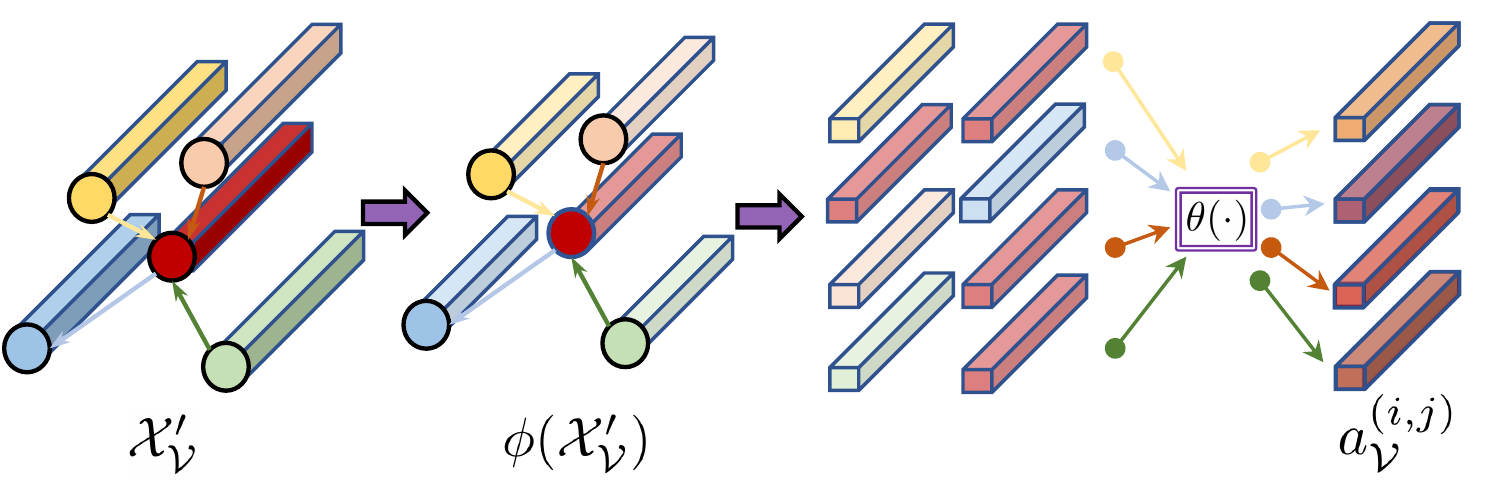}
    \caption{Twinning node attention $\mathcal{A}_{\mathcal{V}}$ design inside our \texttt{EdgeGCN} for modeling the node-driven interactions toward edge evolution.}
    \label{image:Av}
\end{figure}

\noindent\textbf{Twinning node attention.} Similarly, the interactions made by the source-nodes and target-nodes, toward their reaching edges, are modeled by another multi-dimensional attention mask $\mathcal{A}_\mathcal{V} \in \mathcal{R}^{m \times m \times \mathcal{C}^{\prime}_{edge}}$ to be assigned upon $\mathcal{E}_{(\cdot,\cdot)}$, where $\mathcal{C}^{\prime}_{edge}$ is pre-set here as the inner channel number of edge evolution stream to be demonstrated below.
Specifically, given nodes $i$ and $j$, together with the directional edge connecting them $\mathcal{E}_{(i,j)}$, their resulting edge-wise node-driven interaction score $a^{(i,j)}_{\mathcal{V}} \in \mathcal{R}^{1\times1\times\mathcal{C}^{\prime}_{edge}}$ can be learnt from the concatenation of evolved node features that belongs to the sources and targets, as: 
\begin{equation}
a^{(i,j)}_{\mathcal{V}} = 
\sigma \left(
W^T_{\theta}
f (
W^T_{\phi} \mathcal{X}^{\prime}_{\mathcal{V}_i}
\concat
W^T_{\phi} \mathcal{X}^{\prime}_{\mathcal{V}_j}
)
\right),
\label{eq3}
\end{equation}
where $W_{\theta} \in \mathcal{R}^{2\mathcal{C}^{\prime}_{edge}\times\mathcal{C}^{\prime}_{edge}}$ and $W_{\phi} \in \mathcal{R}^{\mathcal{C}_{node}\times\mathcal{C}^{\prime}_{edge}}$ represent the learnable weight matrices of a two-layer structure for transforming channel number from $2\mathcal{C}^{\prime}_{edge}$ to $\mathcal{C}^{\prime}_{edge}$, and from $\mathcal{C}_{node}$ to $\mathcal{C}^{\prime}_{edge}$, respectively.

\noindent\textbf{Edge evolution stream.} With node-driven twinning attention defined, the other evolved edge feature $\mathcal{X}^{\prime}_{\mathcal{V}}$ could now be obtained through:
\begin{equation} \label{eq2}
\mathcal{X}^{\prime}_{\mathcal{E}} = 
f\bigg(W^T_{\mathtt{FC2}}
\Big(
\textcolor{mylayer}{f(W^T_\mathtt{FC1}\mathcal{X}_{\mathcal{E}})}
\odot \mathcal{A}_{\mathcal{V}}
\Big)
\bigg)
,
\end{equation}
where $\mathcal{A}_{\mathcal{V}}$ is the node-driven interactive score, while $W_\mathtt{FC1}\in\mathcal{R}^{\mathcal{C}_{edge} \times\mathcal{C}^{\prime}_{edge}}$ and $W_\mathtt{FC2}\in\mathcal{R}^{\mathcal{C}^{\prime}_{edge} \times\mathcal{C}_{edge}}$ are trainable parameters for two fully-connected layers to transform edge features through $\mathcal{C}_{edge} \xrightarrow{} \mathcal{C}^{\prime}_{edge} \xrightarrow{} \mathcal{C}_{edge}$.

\subsubsection{\textbf{\texttt{EdgeGCN}}}
As illustrated in Fig.~\ref{image:pointsgg}, a joint reasoning module dubbed \texttt{EdgeGCN} capable of exploiting edge features for more comprehensive graph reasoning performed over $\mathcal{G}$ is designed to take as inputs $\mathcal{X}_{\mathcal{V}}$ and $\mathcal{X}_{\mathcal{E}}$ that are initially formed in the Construction$_{\mathbf{SG}}$ stage, conduct collaborative message propagation between two twinning $\mathbf{SG}$ representations enriched by their corresponding edge- and node-driven interactions, and produce the evolved ones to be used by the Inference$_{\mathbf{SG}}$ stage. 
More specifically, \texttt{EdgeGCN} contains two feature evolution streams for the nodes ($\mathcal{X}_{\mathcal{V}} \xrightarrow{} \mathcal{X}^{\prime}_{\mathcal{V}}$) and edges ($\mathcal{X}_{\mathcal{E}} \xrightarrow{} \mathcal{X}^{\prime}_{\mathcal{E}}$), and the evolution stream of each representation is endowed with an attentional interaction mechanism ($\mathcal{A}_\mathcal{E}$ or $\mathcal{A}_\mathcal{V}$) to escort the interdependence between itself and its twinning representation. The detailed architecture designs can be viewed in \textit{Sec.}~\ref{imp}.

As its name suggests, the most distinctive attribute of our \texttt{EdgeGCN} is the explicit modeling of multi-dimensional edge features and their effective interactions with node features for $\mathbf{SG}$ reasoning, compared to other node-wise graph reasoning approaches~\cite{glore,GIN}. 
Noticeably, a \texttt{Vanilla EdgeGCN} without any interactive designs, i.e., $\mathcal{A}_{\mathcal{E}} = \mathcal{A}_{\mathcal{V}} = \mathds{1}$, could be built as two isolated representation learning branches consisting of a two-layer GCN and a two-layer MLP for the independent node and edge evolution.

\subsection{Scene Graph Inference}
\label{sec:3.3}
The final $\mathbf{SG}$ recognition results are predicted on the evolved node features $\mathcal{X}_{\mathcal{V}}^{\prime}$ and edge features $\mathcal{X}_{\mathcal{E}}^{\prime}$. 
Two Multilayer Perceptron based inference streams are established as \texttt{NodeMLP} and \texttt{EdgeMLP} to perform the recognition of objects and their inner structural relationships, respectively. Moreover, \texttt{NodeMLP} and \texttt{EdgeMLP} share the same network structure of two fully-connected layers, but with individual learnable parameters to convert channel numbers through $\mathcal{C}_{in} \xrightarrow{} \frac{\mathcal{C}_{in}}{2} \xrightarrow{} \mathcal{C}_{out}$, where $\mathcal{C}_{in}$ indicates their corresponding input channels and $\mathcal{C}_{out}$ equals to the number of object classes or relationship classes.

\subsection{Implementation Details}
\label{imp}
For the specific instances of $\mathcal{F}_\mathcal{B}(\cdot)$ adopted in Construction$_\mathbf{SG}$ stage, we chose the pioneering PointNet~\cite{pointnet} and its promising follower DGCNN~\cite{dgcnn} for their concise but effective architecture design philosophy, as well as the dynamic and powerful context modeling of each local neighborhood in semantic spaces, respectively. 
Concretely, we set $\mathcal{C}_{input}$ to 9 including 3-dim coordinates, 3-dim RGB colors and 3-dim normal vectors, while $\mathcal{C}_{point}$ set to 256 for unified point-wise feature extraction. 
Within the Reasoning$_\mathbf{SG}$ stage, we set $\mathcal{C}_{node} = 2 \times \mathcal{C}^{\prime}_{node} = 256$, and $\mathcal{C}_{edge} = 2 \times \mathcal{C}^{\prime}_{edge} = 512$ for $\mathbf{SG}$ node and edge evolution streams, respectively, while we used ReLU for $f(\cdot)$ and adopted the Synchronized BatchNorm~\cite{syncbn} for multi-GPU training. 
Regarding the Inference$_{\mathbf{SG}}$ stage, two multi-class cross entropy losses $\mathcal{L}_{node}$ and $\mathcal{L}_{edge}$ were applied for their corresponding $\mathbf{SG}$ representation learning, and hence, the integrated $\mathbf{SGG_{point}}$ framework could be supervised via a joint loss $\mathcal{L}_{\mathbf{SG}} = \mathcal{L}_{node} + \mathcal{L}_{edge}$. 
Please refer to the supplementary materials\footnote{Supplementary materials:~\href{https://sggpoint.github.io/supplementary.pdf}{$\mathbf{SGG_{point}}$.github.io/supplementary.pdf}} for the training details for each dataset.

Before being fed into the Inference$_{\mathbf{SG}}$ stage, the evolved $\mathbf{SG}$ representations obtained from \texttt{EdgeGCN} are combined with the initial ones via a residual connection~\cite{resnet}, to further enhance the discriminative power of graph-based reasoning approaches for $\mathbf{SGG_{point}}$ studies, which is unnecessary for conventional graph representation learning tasks and thus omitted to match up with other GNN setups.

\section{Experiments}

We evaluated the $\mathbf{SGG_{point}}$ framework on both real-world (\textit{Sec.}~\ref{sec:real}) and synthetic 3D scenes (\textit{Sec.}~\ref{sec:Synthetic}), with extensive ablation studies conducted on real-world ones (\textit{Sec.}~\ref{sec:ablation}) to demonstrate the individual contribution of each proposed component toward the overall quality of generated $\mathbf{SG}$s. Despite the studies of $\mathbf{SG}$ representation learning, we also verify the proposed \texttt{EdgeGCN} on five conventional graph representation learning tasks (\textit{Sec.}~\ref{sec:grl}), including three node-wise classification problems and two whole-graph recognition problems.

\subsection{3D $\mathbf{SGG_{point}}$ on Real-World 3D Scans}
\label{sec:real}

\begin{table*}[t]
\begin{tabular}{l c c c c c c } 
\hline
 
\multirow{2}{*}{Graph Reasoning Approach} & \multicolumn{2}{c}{Object Class Prediction} &   \multicolumn{2}{c}{Predicate Class Prediction} & \multicolumn{2}{c}{Relationship Triplet Prediction} \\
           & R@5 & R@10 & F1@3 & F1@5 & R@50 & R@100\\
\hline
$\mathcal{F_B(\cdot)}$ \textit{alone} & \textit{87.40} & \textit{96.26} & \textit{68.55} & \textit{82.79} & \textit{34.97} & \textit{45.86} \\
+ GCN (SGPN)~\cite{3DSSG2020} $^{\ast}$ & 89.61$^{\textcolor{mygreen}{\uparrow}}$ & 96.98$^{\textcolor{mygreen}{\uparrow}}$ & 63.58$^{\textcolor{myred}{\downarrow}}$ & 77.79$^{\textcolor{myred}{\downarrow}}$ & 32.45$^{\textcolor{myred}{\downarrow}}$ &  41.65$^{\textcolor{myred}{\downarrow}}$ \\
+ GloRe$_{\text{PC}}$~\cite{PointCloudGlore}  & 84.06$^{\textcolor{myred}{\downarrow}}$ & 95.17$^{\textcolor{myred}{\downarrow}}$ & 69.23$^{\textcolor{mygreen}{\uparrow}}$  &  80.01$^{\textcolor{myred}{\downarrow}}$ & 31.87$^{\textcolor{myred}{\downarrow}}$ & 42.21$^{\textcolor{myred}{\downarrow}}$\\ 
+ GloRe$_{\text{SG}}$~\cite{glore}  & 85.27$^{\textcolor{myred}{\downarrow}}$ & 96.62$^{\textcolor{mygreen}{\uparrow}}$ & 72.57$^{\textcolor{mygreen}{\uparrow}}$ &  83.42$^{\textcolor{mygreen}{\uparrow}}$ & 29.58$^{\textcolor{myred}{\downarrow}}$ & 38.64$^{\textcolor{myred}{\downarrow}}$ \\

+ \textbf{\texttt{EdgeGCN}} (\textbf{our} $\mathbf{SGG_{point}}$) & \textbf{90.70}$^{\textcolor{mygreen}{\uparrow}}$ & \textbf{97.58}$^{\textcolor{mygreen}{\uparrow}}$ & \textbf{78.88}$^{\textcolor{mygreen}{\uparrow}}$ & \textbf{90.86}$^{\textcolor{mygreen}{\uparrow}}$ & \textbf{39.91}$^{\textcolor{mygreen}{\uparrow}}$ & \textbf{48.68}$^{\textcolor{mygreen}{\uparrow}}$ \\
\hline
\end{tabular}
\caption{Results on real-world 3D scans. Note: ${\ast}$ denotes the usage of two separate $\mathcal{F}_\mathcal{B}(\cdot)$ within Construction$_\mathbf{SG}$ stage, for independent feature extractions of initial node and edge representations in $\mathcal{G}$.}
\label{table:ssg}
\end{table*}

\noindent\textbf{Dataset and evaluation details.} We first validated the effectiveness of our proposed methods on real-world 3D scans using the 3RScan~\cite{Wald2019RIO} dataset.
Extending~\cite{Wald2019RIO} with~\cite{3DSSG2020} results in over one thousand 3D indoor point cloud reconstructions, as well as their corresponding semantic 3D $\mathbf{SG}$ annotations including 27 object classes and 16 relationship categories (details in supplementary materials).
For evaluation, we applied the same scene-level split specified in~\cite{3DSSG2020} on the point cloud representations, which were densely sampled from their released surface reconstructions using CloudCompare~\cite{cloudcompare}, with all mesh information discarded and surface density set as 10k points per square unit.

Following~\cite{IterMessPassing,GraphRCNN,3DSSG2020}, the scene graph prediction performance of the $\mathbf{SGG_{point}}$ framework was evaluated upon the three perspectives, namely object class prediction, predicate class prediction, and relationship triplet prediction. 
More specifically, we adopted the top-k recall metric used in~\cite{visualpriors} for object class prediction and computed the macro-F1 score for predicate class prediction, due to the imbalanced predicate class distribution in~\cite{3DSSG2020}. 
The relationship triplet prediction was jointly generated as an ordered list of (\texttt{subject}, \texttt{predicate}, \texttt{object}) triplets, whose triplet-level confidence scores were obtained by multiplying each respective score~\cite{GraphRCNN}, and the most confident ones are separated for evaluation against the ground truth annotations~\cite{3DSSG2020}.

\noindent\textbf{Experimental results.} We first set the $\mathcal{F}_\mathcal{B}(\cdot)$ \textit{alone} baseline without invoking any graph reasoning modules. 
Note: $\mathcal{F}_\mathcal{B}(\cdot)$ was implemented as PointNet~\cite{pointnet} here to make fair comparisons with the current benchmark~\cite{3DSSG2020} and justify various graph reasoning effects, while we also demonstrated that our method could be further enhanced consistently by changing the backbone in the coming ablation studies.

\begin{figure*}[t]
    \centering
    \includegraphics[width=\textwidth,height=0.25\textwidth]{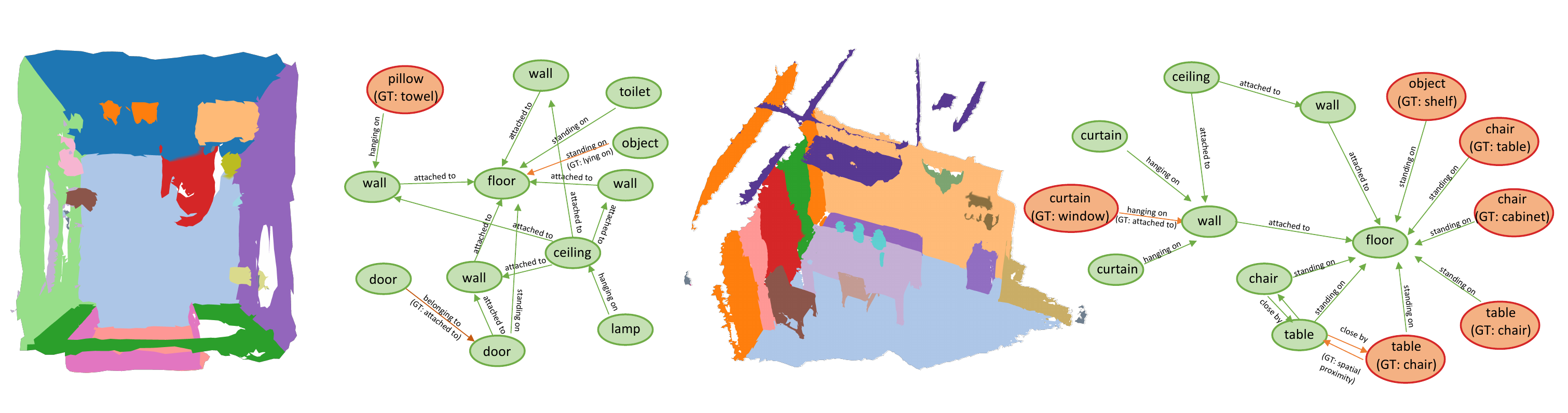}
    \caption{Qualitative analysis of the $\mathbf{SGG_{point}}$ framework. For visualization purpose, misclassified object or structural relationship predictions are indicated with ground truth (GT) values in red, while the correct ones are shown in green with GT values omitted.}
    \label{image:demo}
    \vspace{-3mm}
\end{figure*}

We then reproduced SGPN~\cite{3DSSG2020}, which was similar to~\cite{ImgGeneration_CVPR18} that treated both objects and their interrelationships as graph nodes to conduct message propagation with GCN~\cite{GCN}, to generate the acquired triplets. 
As shown in Table~\ref{table:ssg}, employing GCN for scene graph reasoning in an intuitive way as presented in~\cite{3DSSG2020} could improve the object class recognition but may harm the performance in other two $\mathbf{SG}$ tasks, which confirms the empirical findings reported in~\cite{3DSSG2020} on over-smoothing issue caused by multi-layer GCNs. 
We next verified GloRe modules to perform global relation reasoning on scene graph representation learning using official implementations in their released repository~\cite{glore}. 
The GloRe module could be applied at various positions to reach reasoning effects at two different levels, namely the point cloud level (GloRe$_{\text{PC}}$)~\cite{PointCloudGlore} and scene graph level (GloRe$_{\text{SG}}$)~\cite{glore}.
In contrast to SGPN, both GloRe$_{\text{PC}}$ and GloRe$_{\text{SG}}$ tend to benefit the predicate class prediction and may damage the recognition under two other metrics as a trade-off.
Our \texttt{EdgeGCN} achieved the best results under all evaluation metrics, which demonstrated the superiority of edge-oriented relationship modeling, as well as its associated twinning attentions between graph nodes and edges, for scene graph reasoning. The qualitative visualization can be viewed in Fig.~\ref{image:demo}.

\subsection{Ablation Studies}
\label{sec:ablation}

\begin{table}[b]
\tabcolsep=0.11cm
\begin{tabular}{|l?l|l|l|l?l|l|} 
\hline
\multirow{2}{*}{ID} &
\multirow{2}{*}{Task} &
  \multicolumn{3}{c?}{Reasoning$_\mathbf{SG}$} &
  \multicolumn{2}{c|}{$\mathbf{SG}$ Recognition} \\\cline{3-7} 
  
\multirow{2}{*}{} & \multirow{2}{*}{}   & GNNs & $\mathcal{A}_\mathcal{E}$ & $\mathcal{A}_\mathcal{V}$  &  node $_\mathtt{R@1}$ &  edge $_\mathtt{F1@1}$ \\
\hline
\hline
\texttt{A} w/ $\diamond$   &  N & - & - & - & 48.6 & -   \\ 
\texttt{B} w/ $\diamond$  &  N & GCN & - & - & 54.4 & -   \\ 
\texttt{C} w/ $\diamond$  &  N+E & - & - & - & 48.4&  38.7\\ 
\hline
\texttt{D} w/ $\diamond$  &  N+E & \texttt{EdgeGCN} & $\times$ & $\times$ & 54.8 \large{\textcircled{\small{4}}}&  41.9 \large{\textcircled{\small{3}}}\\ 
\texttt{E} w/ $\diamond$  &  N+E & \texttt{EdgeGCN} & $\surd$ & $\times$ & 56.9 \large{\textcircled{\small{2}}}& 41.1 \large{\textcircled{\small{4}}}\\ 
\texttt{F} w/ $\diamond$  &  N+E & \texttt{EdgeGCN} & $\times$ & $\surd$ & 56.4 \large{\textcircled{\small{3}}}&  \textbf{50.0 \large{\textcircled{\small{1}}}}\\ 
\textbf{\texttt{G} w/ $\diamond$ } &  \textbf{N+E} & \textbf{\texttt{EdgeGCN}} & $\surd$ & $\surd$ &\textbf{57.1 \large{\textcircled{\small{1}}}}& 48.7 \large{\textcircled{\small{2}}}\\ 
\hline

\hline
\texttt{A} w/ $\star$   &  N & - & - & - & 58.0 & -   \\ 
\texttt{B} w/ $\star$  &  N & GCN & - & - & 61.3 & -   \\ 
\texttt{C} w/ $\star$  &  N+E & - & - & - & 57.1&39.6\\ 
\hline
\texttt{D} w/ $\star$  &  N+E & \texttt{EdgeGCN} & $\times$ & $\times$ &60.8 \large{\textcircled{\small{4}}}& 43.9 \large{\textcircled{\small{3}}}\\ 
\texttt{E} w/ $\star$  &  N+E & \texttt{EdgeGCN} & $\surd$ & $\times$ &61.7 \large{\textcircled{\small{2}}}& 41.1 \large{\textcircled{\small{4}}}\\ 
\texttt{F} w/ $\star$  &  N+E & \texttt{EdgeGCN} & $\times$ & $\surd$ & 61.0 \large{\textcircled{\small{3}}} & 47.4 \large{\textcircled{\small{2}}}  \\ 
\textbf{\texttt{G} w/ $\star$}  &  \textbf{N+E} & \textbf{\texttt{EdgeGCN}} & $\surd$ & $\surd$ & \textbf{62.5 \large{\textcircled{\small{1}}}} & \textbf{49.7 \large{\textcircled{\small{1}}}}\\ 
\hline
\end{tabular}
\caption{Ablation studies of $\mathbf{SGG_{point}}$ framework. Task (N) and Task (E) represent the node and edge recognition tasks for $\mathbf{SGG_{point}}$ studies, respectively, while dash lines indicate 'not applicable'. Two specific $\mathcal{F}_\mathcal{B}(\cdot)$ implementations include PointNet ($\diamond$) and DGCNN ($\star$). \textcircled{$k$} denotes rankings within each sub-block.} 
\label{table:edgegcn_ablation}
\end{table}

To reveal the precise performance gain of each proposed component, we instead reported the respective recognition results of objects and relationships in top-1 manner within our ablation studies. 
\texttt{Models A-C} were raised to establish the baselines of the $\mathbf{SGG_{point}}$ framework, especially for the Construction$_\mathbf{SG}$ and Inference$_\mathbf{SG}$ stages, while \texttt{models D-G} were built to demonstrate the effectiveness of adopting multi-dimensional edge features $\mathcal{X}_{\mathcal{E}}$ for scene graph reasoning, together with two associated twinning interactions between nodes and edges. 

\noindent\textbf{Model designs.} More specifically, \texttt{model A} indicates the baseline performance of utilizing backbone networks for object classification in the $\mathbf{SG}$ context, leaving the initialization of $\mathcal{X}_{\mathcal{E}}$, edge evolution stream, and \texttt{EdgeMLP} untouched in Construction$_\mathbf{SG}$, Reasoning$_\mathbf{SG}$, and Inference$_\mathbf{SG}$ stages, respectively. 
\texttt{Model B} enhances \texttt{model A} via the same two-layer GCN adopted in our \texttt{EdgeGCN} to perform a $\mathcal{X}_{\mathcal{E}}$-uninvolved graph reasoning, which relies on the node information alone, where as \texttt{model C} enriches \texttt{model A} to be compatible with the joint recognition tasks for $\mathbf{SG}$ objects and relationships via adding edge-wise supervisions into the model training. 
Furthermore, a \texttt{Vanilla EdgeGCN} is trained, as \texttt{model D}, to fulfill \texttt{model B} with the necessary components for relationship recognition. 
Then, two twinning interactive attention schemes are equipped to \texttt{model D}, investigating the independent impacts of edge-driven and node-driven interactions from \texttt{model E} and \texttt{F}, respectively. 
At the end, with the aim of reaching a comprehensive $\mathcal{X}_{\mathcal{E}}$-involved $\mathbf{SG}$ reasoning effect, \texttt{model G} is constructed with collaborative improvements invoked from both node and edge sides for $\mathbf{SG}$ representation evolution.

\noindent\textbf{Discussions.} As details revealed in Table~\ref{table:edgegcn_ablation}, graph reasoning could benefit the object recognition process (\texttt{A}$^{\diamond/\star}\xrightarrow{}$\texttt{B}$^{\diamond/\star}$), while roughly adding edge-wise supervision may harm the outcomes by contrast (\texttt{A}$^{\diamond/\star}\xrightarrow{}$\texttt{C}$^{\diamond/\star}$) and we attribute it as the potential distractions on the unified extraction of point-wise features via $\mathcal{F}_\mathcal{B}(\cdot)$. 
These distractions could be partially alleviated by adding extra trainable parameters such that more degree of freedom might be positively introduced into the edge representation learning (\texttt{B}$^{\diamond}\xrightarrow{}$\texttt{D}$^{\diamond}$; \texttt{C}$^{\diamond/\star}\xrightarrow{}$\texttt{D}$^{\diamond/\star}$), which thus forms the edge evolution stream in our \texttt{Vanilla EdgeGCN}. 
The effectiveness of $\mathcal{A}_\mathcal{E}$ could be somewhat controversial, as it brings improvements on node recognition results yet reduces the edge ones as a trade-off (\texttt{D}$^{\diamond/\star}\xrightarrow{}$\texttt{E}$^{\diamond/\star}$), and we blame it as similar distractions discussed above. The other twinning interaction module $\mathcal{A}_\mathcal{E}$ could accelerate the edge evolution as expected, without distressing the edge evolution (\texttt{D}$^{\diamond/\star}\xrightarrow{}$\texttt{F}$^{\diamond/\star}$). Stunningly, combining $\mathcal{A}_\mathcal{E}$ and $\mathcal{A}_\mathcal{V}$ not only achieves the best performance in terms of object classifications (\texttt{\{E,F\}}$^{\diamond/\star}\xrightarrow{}$\texttt{G}$^{\diamond/\star}$), but also practically curbs the distractions made by $\mathcal{A}_\mathcal{E}$ (\texttt{E}$^{\diamond}\xrightarrow{}$\texttt{G}$^{\diamond}$) and even produces the best outcomes (\texttt{E}$^{\star}\xrightarrow{}$\texttt{G}$^{\star}$) in terms of relationship predictions.

\subsection{3D $\mathbf{SGG_{point}}$ on Synthetic 3D Scenes}
\label{sec:Synthetic}

\begin{table}[b]
\tabcolsep=0.17cm
\begin{tabular}{l c c c c} 
\hline
Method & Bed & Living & Bath & Office \\
\hline
GRAINs~\cite{li2018grains} $^{\ast}$ & 45.1 & 43.7 & 42.4 & 45.6  \\
Wang et al.~\cite{Kai1} $^{\ast}$ & 48.9 & 46.6 & 61.4 & 46.6  \\
MVCNN~\cite{multiview} &  69.6 & 55.8 & 43.4 & 67.8 \\ 
SceneGraphNet~\cite{zhou2019scenegraphnet} $^{\ast}$ & 66.8 & 67.6 & \textbf{69.8} & 64.8 \\
SceneGraphNet~\cite{zhou2019scenegraphnet}  &  \textbf{79.9} & 74.7 & 56.4 & 73.0 \\ 
$\mathbf{SGG_{point}}$ \textbf{w/} $\diamond$ & 79.5 & \textbf{76.2} & 61.6 & \textbf{74.1} \\
\hline

\end{tabular}
\caption{Results on synthetic 3D scenes, where ${\ast}$ denotes missing object predictions achieved by scene synthesis based approaches.}
\label{table:suncg}
\vspace{-4mm}
\end{table}

\noindent\textbf{Dataset and evaluation details.} We further analyzed our methods on the SUNCG~\cite{song2016suncg} dataset, to verify its generalization ability on synthetic 3D scenes. 
The SUNCG dataset is comprised of over 45k 3D virtual scenes, which were manually created with the Planner5d platform~\cite{planner5d} in four room categories, i.e., office, bedroom, bathroom, and living room. 
As suggested in~\cite{zhou2019scenegraphnet}, we filtered out the non-rectangular scenes to maintain fair comparisons with previous studies~\cite{li2018grains, Kai1, zhou2019scenegraphnet}, then repeated the whole experimental procedure independently for each synthetic room category~\cite{Kai1,kai2}, as each category owns unique objects.

For evaluation, we followed the same dataset splitting and preprocessing policy in~\cite{zhou2019scenegraphnet} and adopted three-class inter-object relationship annotations (support, surround, and next-to)~\cite{zhou2019scenegraphnet,li2018grains} as the predicate ground truth to train $\mathbf{SGG_{point}}$ on synthetic 3D scenes, with point cloud sampling settings unchanged to previous studies. The $\mathbf{SG}$ object classification accuracy was reported to compare $\mathbf{SGG_{point}}$ with other existing SOTAs on SUNCG dataset.

\noindent\textbf{Results and discussions.} 
Some early methods~\cite{li2018grains,Kai1} on this dataset were merely designed for scene synthesis studies, where they firstly removed the target objects from the scenes and then utilized their surrounding contexts to predict the missing labels for scene synthesis evaluations. 
We thus recognized their approaches as \textit{missing} object prediction studies and included their results as reference benchmarks (in Table~\ref{table:suncg}) for comparisons with traditional object recognition methods such as~\cite{multiview} and~\cite{zhou2019scenegraphnet}. 
Unlike these missing object predictions who treat target objects as empty nodes and take as inputs the contextual scene formed by other available nodes in $\mathcal{G}$, the traditional ones including ours would instead extract visual features from target objects themselves for further usage (e.g., \texttt{EdgeGCN}). 
As shown in Table~\ref{table:suncg}, our $\mathbf{SGG_{point}}$ outperformed existing SOTAs on \textit{Living} and \textit{Office} datasets, and we achieved on-par result on \textit{Bed} category.
In contrast to~\cite{zhou2019scenegraphnet}, our \texttt{EdgeGCN} employed multi-dimensional edge features and it is thus insensitive to specific types of inter-object relationships.
Besides, compared to the multi-view based approaches~\cite{multiview,zhou2019scenegraphnet}, $\mathbf{SGG_{point}}$ supports direct point-wise manipulations on 3D scenes via a more efficient manner, in terms of the time and space complexity~\cite{pointnet}.

\subsection{Graph Representation Learning}
\label{sec:grl}
The effectiveness of our \texttt{EdgeGCN} could also be verified on graph representation learning studies, such as node-wise classification and whole-graph recognition problems. 
More specifically, our method was evaluated on three popular citation network datasets (Cora, CiteSeer, and Pubmed)~\cite{Planetoid} and two molecular datasets (Tox21 and BBBP)~\cite{molecular_datasets}.

\begin{table}[b]
\tabcolsep=0.03cm
\small
\begin{tabular}{|c?c|c|c?c|c|} 
\hline
\multirow{2}{*}{GNNs} &  \multicolumn{3}{c?}{Node $_{\mathtt{Accu.}}$} & \multicolumn{2}{c|}{Graph $_\mathtt{AUC}$} \\\cline{2-6} 
 & Cora  &CiteSeer & Pubmed & Tox21 & BBBP \\
\hline
\hline
GCN  &  \textcolor{mygray}{80.3$_{\pm  0.7}$} & \textcolor{mygray}{67.7$_{\pm  0.8}$} & \textcolor{mygray}{78.5$_{\pm  0.5}$} & \textcolor{mygray}{73.1$_{\pm  1.1}$} &
\textcolor{mygray}{64.3$_{\pm  3.5}$}\\ 
\hline
GAT (8, 4)  & 79.8$_{\pm  0.8}$ & 68.1$^{\textcolor{mygreen}{+0.4\uparrow}}_{\pm  0.9}$  & 76.7$_{\pm  0.7}$  & \multirow{4}{*}{68.3$_{\pm  1.8}$  }& \multirow{4}{*}{\textbf{65.1$^{\textcolor{mygreen}{\textbf{+0.8}\uparrow}}_{\pm 1.9}$ }}\\ 
GAT (8, 8)  & 79.5$_{\pm  0.7}$ & 68.0$_{\pm  1.1}$ & 76.4$_{\pm  0.8}$ & &\\ 
GAT (16, 4) & 79.7$_{\pm  1.0}$ & 67.9$_{\pm  1.0}$ & 76.4$_{\pm  0.9}$ & &\\ 
GAT (16, 8) & 79.8$_{\pm  1.0}$ & 67.5$_{\pm  1.6}$ & 76.1$_{\pm  1.1}$ & &\\ 
\hline
EGNN(A)     & 81.1$^{\textcolor{mygreen}{+0.8\uparrow}}_{\pm  0.7}$ & 68.5$^{\textcolor{mygreen}{+0.8\uparrow}}_{\pm  0.8}$ & 79.4$_{\pm  0.5}$ & 73.3$^{\textcolor{mygreen}{+0.2\uparrow}}_{\pm  1.2}$ &64.5$^{\textcolor{mygreen}{+0.2\uparrow}}_{\pm  3.1}$ \\ 
EGNN(C)     & 80.9$_{\pm  0.7}$ & 67.9$_{\pm  0.6}$ & \textbf{79.5$^{\textcolor{mygreen}{\textbf{+1.0}\uparrow}}_{\pm  0.4}$ } & 73.2$^{\textcolor{mygreen}{+0.1\uparrow}}_{\pm  1.2}$ & 63.9$_{\pm  2.9}$\\ 
\hline
\textbf{\texttt{EdgeGCN}}   & \textbf{81.6$^{\textcolor{mygreen}{\textbf{+1.3}\uparrow}}_{\pm  0.7}$} & \textbf{69.4$^{\textcolor{mygreen}{\textbf{+1.7}\uparrow}}_{\pm  0.9}$ } & 78.7$^{\textcolor{mygreen}{+0.2\uparrow}}_{\pm  0.4}$ & \textbf{73.7$^{\textcolor{mygreen}{\textbf{+0.6}\uparrow}}_{\pm  0.9}$}&64.6$^{\textcolor{mygreen}{+0.3\uparrow}}_{\pm  3.8}$\\ 
\hline

\end{tabular}
\caption{Conventional graph representation learning tasks, including node classification on citation network datasets and graph recognition for molecular analysis.}
\label{table:node_classification_task}
\vspace{-4mm}
\end{table}

Since these conventional graph representation learning tasks do not provide edge annotation, we thus omitted our edge evolution stream, together with its associated twinning node attention mechanism, and compared the resulting \texttt{EdgeGCN} ($\mathcal{A}_\mathcal{E}$) with its counterparts including GCN~\cite{GCN}, GAT~\cite{GAT}, and EGNNs, i.e., EGNN(A) and EGNN(C), which were reproduced in accordance to their reported settings~\cite{EGNN}. 
We applied a Pytorch Geometric~\cite{pyg} script and a DGL~\cite{dgl} script, for evaluations conducted on citation network datasets and molecular datasets, respectively. We kept all specific training settings unchanged for all method evaluations, except for repeating their procedure 50 times for each approach and reporting the averaged accuracy (Accu.), or area under the ROC curve (AUC), with standard deviation to reach reliable comparisons.

\noindent\textbf{Node-wise classification for citation analysis.}
The GCN, GAT, and our \texttt{EdgeGCN} ($\mathcal{A}_\mathcal{E}$) were constructed as two-layer networks. 
Since~\cite{pyg} does not provide a universal GAT implementation, we reproduced GAT with various settings according to~\cite{GAT}, where GAT ($C_{inner}$, $K$) denotes its network settings, i.e., $C_{inner}$ inner channels and $K$ attention heads. $C_{inner}$ was set by default to 16 for all other graph networks.
As shown in Table~\ref{table:node_classification_task}, our \texttt{EdgeGCN} ($\mathcal{A}_\mathcal{E}$) has shown its superior performance over all competitors under the same training settings on Cora and CiteSeer, and achieved on-par result on Pubmed. 
Unlike other designs such as GAT and EGNN, our method does not rely on the hyper-parameter settings or different types of layer instance.

\begin{figure}[t]\raggedright
\includegraphics[width=\columnwidth]{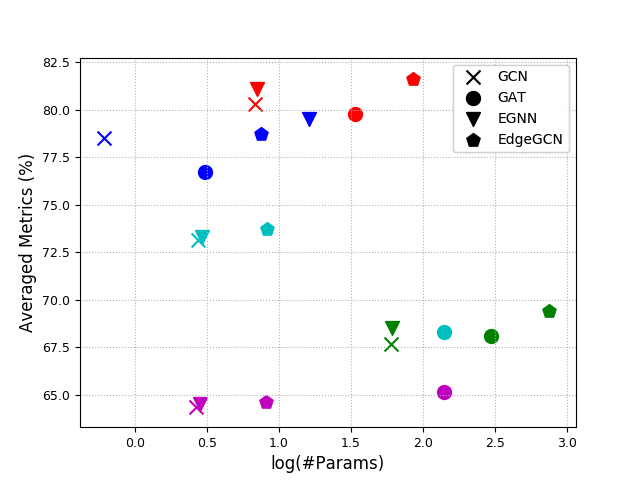}
\caption{GNN comparisons on various graph representation learning tasks on \textcolor{cora}{\textit{Cora}}, \textcolor{citeseer}{\textit{CiteSeer}}, \textcolor{pubmed}{\textit{Pubmed}}, \textcolor{tox21}{\textit{Tox21}}, and \textcolor{bbbp}{\textit{BBBP}} datasets. \texttt{\#Params} denotes the number of trainable parameters (K).}
\label{image:grl}
\vspace{-4mm}
\end{figure}

\noindent\textbf{Whole-graph recognition for molecular analysis.}
We adopted the universal three-layer instances of GCN and GAT provided by~\cite{dgl} and extended our \texttt{EdgeGCN} to three layers as well, with $\mathcal{A}_{\mathcal{E}}$ inserted to the second layer. As shown in Table~\ref{table:node_classification_task}, the significance of our \texttt{EdgeGCN}($\mathcal{A}_\mathcal{E}$) design could be verified on both Tox21 and BBBP datasets under the same evaluation protocol applied. Fig.~\ref{image:grl} demonstrates the trade-off between effectiveness and efficiency.

\section{Conclusion}

To endow GCNs with edge-assisted reasoning capability, we introduced an edge-oriented GCN dubbed \texttt{EdgeGCN} to learn a pair of twinning interactions between nodes and edges, so that comprehensive $\mathbf{SG}$ reasoning could thus be conducted to enhance each individual evolution. 
Taking \texttt{EdgeGCN} as the core component, we proposed an integrated $\mathbf{SGG_{point}}$ framework to tackle 3D point-based scene graph generation problems through three sequential stages. 
Overall, our integrated $\mathbf{SGG_{point}}$ framework was established to seek and infer scene structures of interest from both real-world and synthetic 3D point-based scenes. 
Moreover, we also validated our edge-driven reasoning scheme on conventional graph representation learning benchmark datasets for citation network and molecular analysis.

{\small
\bibliographystyle{ieee_fullname.bst}
\bibliography{cvpr.bib}
}

\end{document}